\documentclass[11pt]{article}

\usepackage[utf8]{inputenc}
\usepackage[T1]{fontenc}
\usepackage{amsmath,amssymb,amsthm}
\usepackage{mathtools}
\usepackage{bm}
\usepackage{graphicx}
\usepackage{booktabs}
\usepackage{hyperref}
\usepackage{url}
\usepackage{natbib}
\usepackage{algorithm}
\usepackage{algorithmic}
\usepackage{caption}
\usepackage{subcaption}
\usepackage{enumitem}
\usepackage{xcolor}
\usepackage[margin=1in]{geometry}

\hypersetup{
    colorlinks=true,
    linkcolor=blue,
    filecolor=magenta,      
    urlcolor=cyan,
    citecolor=blue,
    pdftitle={GO-OSC and VASH: Geometry-Aware Representation Learning for Early Degradation Detection in Oscillatory Systems},
    pdfauthor={Vashista Nobaub},
}

\newtheorem{theorem}{Theorem}[section]
\newtheorem{proposition}[theorem]{Proposition}
\newtheorem{lemma}[theorem]{Lemma}

\newtheorem{definition}[theorem]{Definition}
\newtheorem{assumption}[theorem]{Assumption}
\newtheorem{remark}[theorem]{Remark}

\newcommand{\R}{\mathbb{R}}
\newcommand{\Z}{\mathbb{Z}}
\newcommand{\N}{\mathcal{N}}
\newcommand{\E}{\mathbb{E}}
\newcommand{\Var}{\mathrm{Var}}
\newcommand{\argmin}{\operatorname{argmin}}

\newcommand{\Thetacan}{\Theta_{\mathrm{can}}}
\newcommand{\Vgeo}{\mathcal{V}_{\mathrm{Geo}}}
\newcommand{\Ven}{\mathcal{V}_{\mathrm{En}}}

\title{\textbf{GO-OSC and VASH: Geometry-Aware Representation Learning for Early Degradation Detection}}

\author{
Vashista Nobaub\\
Datar Consulting\\
\texttt{labs@datar.fr}
}

\date{}

\begin{document}

\maketitle

\begin{abstract}
Early-stage degradation in oscillatory systems typically appears as \textit{geometric distortions} of the dynamics---such as phase jitter, frequency drift, or loss of coherence---well before changes in signal energy are detectable. In this regime, classical energy-based diagnostics and unconstrained learned embeddings are structurally insensitive, leading to late or unstable detection. We introduce \textbf{GO-OSC}, a geometry-aware representation learning framework for oscillatory time series that enforces a \textbf{canonical and identifiable latent parameterization}, enabling stable comparison and aggregation across short, unlabeled windows. Building on this representation, we define a family of invariant \textbf{linear geometric probes} of degradation-relevant directions in latent space. We provide theory showing that under early phase-only degradation, energy-based statistics have zero first-order detection power, whereas geometric probes achieve strictly positive sensitivity. Our main contribution is a precise characterization of \textit{when and why linear probing fails under non-identifiable representations}, and a principled solution via canonicalization. Experiments on synthetic benchmarks and real vibration datasets validate the theory, demonstrating earlier detection, improved data efficiency, and robustness across operating regimes.
\end{abstract}

\noindent\textbf{Code availability:} All code is available at \url{https://github.com/DatarConsulting/GO-OSC}

\section{Introduction}

Early-stage degradation in many physical and cyber--physical systems is not energetic---it is geometric. In oscillatory dynamics such as rotating machinery, power grids, and biological rhythms, incipient faults often manifest first as subtle distortions of phase geometry: jitter, frequency wander, or loss of coherence. During this regime, signal energy and second-order statistics remain essentially unchanged, rendering classical diagnostics (e.g., RMS, spectral power, kurtosis) and many learned features ineffective until degradation is well advanced \citep{jardine2006review, randall2011vibration, basseville1993detection}.

\vspace{0.5em}
\noindent\fbox{\parbox{0.97\textwidth}{
\textbf{Failure Mode Studied.} This paper studies a specific but fundamental regime: \textit{phase-only degradation under invariant second-order statistics}. In this regime, any representation whose latent coordinates are not identifiable across windows renders linear probing statistically ill-defined, independent of model capacity. We characterize when and why this failure occurs, and show how canonicalization restores statistical detectability.
}}
\vspace{0.5em}

This observation raises a representation learning question that is rarely made explicit:

\begin{quote}
\textit{What representation makes early, phase-only degradation linearly detectable from short, unlabeled time windows?}
\end{quote}

A large body of modern time-series representation learning---contrastive, predictive, or self-supervised---aims to learn generic embeddings that support a wide range of downstream tasks via linear probing \citep{bengio2013representation, lecun2015deep, franceschi2019unsupervised}. In practice, however, such embeddings are typically \textit{unconstrained and non-identifiable}: latent coordinates are defined only up to arbitrary rotations, scalings, and permutations. While this flexibility is often benign in classification settings, it becomes a structural limitation for early degradation detection, where signals are weak, labels are scarce, and statistics must be aggregated consistently across windows and operating regimes.

In this regime, the problem is not insufficient model capacity or training data. Rather, it is that without identifiability, linear probes are statistically ill-defined: the same physical state can map to incompatible latent coordinates across windows, masking subtle but systematic geometric changes.

This paper argues that early degradation detection is fundamentally a \textit{geometry-aware representation learning problem}. To address it, we introduce GO-OSC, a structured oscillatory representation that enforces a canonical and identifiable latent parameterization. By restricting latent dynamics to a canonical real--Schur oscillatory gauge, GO-OSC resolves the similarity-transform ambiguities inherent in latent state-space and learned dynamical models \citep{ljung1999system, kailath1980linear}. The resulting latent coordinates are unique (up to measure-zero degeneracies), stable across windows, and directly comparable across operating conditions.

Crucially, this canonicalization is not a modeling convenience---it is what makes principled linear probing possible. Building on GO-OSC, we define a family of \textbf{linear geometric probes}: simple, invariant geometric functionals that probe degradation-relevant directions in latent space. Each probe corresponds to a specific geometric failure mode (e.g., phase coherence collapse, frequency wander) and is statistically well-defined under aggregation.

We provide intuition-first theory explaining why this design matters. Under early phase-only degradation, energy-based statistics---and more generally, statistics depending only on second moments---are second-order insensitive: they have zero first-order detection power. In contrast, geometric probes derived from a canonical representation respond at first order. We formalize this separation under a local asymptotic normality (LAN) regime, showing that geometric probes achieve strictly positive Pitman efficiency while energy-based statistics do not \citep{vaart1998asymptotic, lecam2000asymptotics}.

Empirical results on synthetic benchmarks and real vibration datasets confirm this analysis. GO-OSC combined with linear geometric probes detects degradation earlier, with substantially improved data efficiency and robustness to operating condition changes compared to energy-based diagnostics and unconstrained learned representations. Ablation studies demonstrate that canonicalization is essential: without it, geometric probes become unstable and lose detection power.

\textit{While our empirical evaluation focuses on oscillatory signals, the contribution of this paper is not a new benchmark, but a characterization of when and why representation learning fails under weak geometric perturbations.}

This work provides a general blueprint for structured representation learning in physical systems \citep{bronstein2021geometric}:
\begin{enumerate}[label=(\arabic*)]
    \item make task-relevant geometry explicit in the latent space,
    \item enforce canonical structure to ensure identifiability and comparability,
    \item probe degradation directions with invariant, statistically grounded linear functionals.
\end{enumerate}

This perspective reframes early fault detection not as feature engineering or classification, but as the design of representations in which subtle, safety-critical signals become linearly and reliably observable.

\subsection{Contributions}

\begin{enumerate}
    \item \textbf{GO-OSC: Geometry-Aware Oscillatory Representation.}
    We introduce GO-OSC, a structured representation learning method for oscillatory time series that enforces a canonical real--Schur gauge. This resolves similarity-transform non-identifiability, yielding unique, interpretable, and stable latent coordinates that are comparable across windows and operating regimes.

    \item \textbf{Linear Geometric Probes.}
    We define a family of gauge-invariant geometric functionals that act as linear probes of degradation-relevant directions in latent space. Each probe is interpretable, statistically well-defined, and aligned with a specific geometric failure mode (e.g., phase coherence collapse, frequency wander).

    \item \textbf{Theory: Why Geometry Beats Energy.}
    We provide theoretical results showing that, under early phase-only degradation, energy-based statistics have zero first-order detection power, whereas geometric probes have strictly positive power. We prove that the optimal linear test statistic can be constructed from our geometric probes, and formalize this separation under a local asymptotic normality regime.

    \item \textbf{Empirical Validation.}
    Experiments on synthetic benchmarks and real vibration datasets demonstrate that GO-OSC with linear geometric probes detects degradation earlier than energy-based and generic learned representations, with substantially improved data efficiency and robustness to operating condition changes.

    \item \textbf{A Blueprint for Structured Representation Learning.}
    Beyond vibration monitoring, this work illustrates how enforcing canonical structure and geometric invariance can yield representations that are not only predictive, but also identifiable, stable, and theoretically analyzable---key desiderata for ML in physical systems.
\end{enumerate}

\paragraph{Negative Result.} Our analysis implies that generic self-supervised representations trained without identifiability constraints cannot reliably support linear probing for early phase-only degradation, even if they achieve strong performance on downstream classification tasks. This limitation is architectural, not statistical: no amount of data or model capacity can fix linear probing when representations are non-identifiable across aggregation windows.

\section{Related Work}

This work sits at the intersection of time-series representation learning, structured latent variable models, and fault detection in physical systems. We highlight connections to these areas and clarify how our approach differs.

\paragraph{Time-Series Representation Learning.}
Modern representation learning for time series emphasizes self-supervised and contrastive objectives that produce generic embeddings useful across downstream tasks \citep{bengio2013representation, lecun2015deep}. Contrastive predictive coding \citep{oord2018representation} and its variants learn representations by predicting future latent states, while recent methods like TS2Vec \citep{yue2022ts2vec} achieve strong performance through hierarchical contrastive learning. A comprehensive survey of self-supervised approaches for time series is provided by \citet{zhang2024self}. Earlier work on multivariate time series learned representations through temporal contrast and transformation invariance \citep{franceschi2019unsupervised}. While effective in data-rich settings, these approaches typically learn \textit{unconstrained latent spaces} that lack identifiability and may entangle amplitude, phase, and frequency information. As a result, they often require substantial data to detect subtle, early-stage changes that are localized to specific geometric aspects of the dynamics.

In contrast, GO-OSC imposes an explicit oscillatory inductive bias and enforces a canonical latent parameterization. Rather than learning arbitrary embeddings, it restricts representations to a geometry-aware hypothesis class in which degradation-relevant directions are explicitly represented and linearly accessible. Our linear geometric probes can be viewed as task-aligned probes that directly exploit this structure.

\paragraph{Latent State-Space Models and Structured Representations.}
Latent state-space models provide a principled framework for modeling temporal dynamics, with foundations in optimal filtering and control theory \citep{kalman1960new, anderson1979optimal}. System identification methods have established conditions for consistent parameter estimation \citep{ljung1999system}, and linear systems theory provides tools for analyzing observability and controllability \citep{kailath1980linear}. However, latent state representations are generally identifiable only up to similarity transforms, leading to instability and non-comparability across training runs or time windows.

Several works address identifiability through architectural constraints or normalization schemes, but few provide a canonical parameterization with formal guarantees. GO-OSC resolves this issue by restricting parameters to a canonical real--Schur oscillatory gauge, leveraging classical results on matrix canonical forms \citep{horn2012matrix, golub2013matrix}. This canonicalization is essential for producing stable representations that support aggregation, statistical analysis, and consistent downstream probing.

\paragraph{Inductive Bias and Geometry in Representation Learning.}
Incorporating domain structure and geometric priors into representation learning has been shown to improve sample efficiency, robustness, and interpretability, particularly in scientific and physical domains. The emerging field of geometric deep learning provides a unifying framework for understanding how symmetries and invariances can be encoded in neural architectures \citep{bronstein2021geometric}. Group-equivariant networks demonstrate how respecting symmetry structure improves generalization \citep{cohen2016group}.

Our approach aligns with this line of work by making \textit{oscillatory geometry} explicit in the latent space. Rather than enforcing invariance through data augmentation or contrastive objectives, GO-OSC embeds geometric structure directly into the representation via canonicalization, enabling invariant, interpretable functionals (VASH indicators) to be computed reliably.

\paragraph{Fault Detection and Early Degradation Analysis.}
Condition monitoring and prognostics for machinery systems has a long history, with comprehensive reviews covering classical and modern approaches \citep{jardine2006review}. Vibration-based methods form a cornerstone of practical diagnostics \citep{randall2011vibration}. The underlying signal analysis draws on spectral methods \citep{priestley1981spectral} and time-frequency representations \citep{cohen1995time}. For oscillatory systems, phase coherence and synchronization phenomena provide important diagnostic signatures \citep{pikovsky2001synchronization}. Recent machine learning approaches to fault diagnosis are surveyed by \citet{lei2020applications}, covering deep learning and transfer learning methods.

A key challenge is detecting abrupt changes before they cause system failure \citep{basseville1993detection}. Classical energy-based statistics respond late when degradation initially affects phase or coherence rather than amplitude. More recent learning-based approaches apply generic classifiers or embeddings to raw signals, often achieving strong performance in supervised regimes but offering limited interpretability and few guarantees under distribution shift.

This work departs from both approaches by focusing explicitly on \textit{early, phase-only degradation} and by explaining failure modes of energy-based and unconstrained learned representations through local asymptotic analysis \citep{vaart1998asymptotic, ibragimov1981statistical}. The resulting geometric indicators are not heuristic features but statistically grounded probes derived from an identifiable representation.

\paragraph{Summary of Differences.}
In contrast to prior work, this paper:
\begin{itemize}
    \item Treats early degradation detection as a \textit{representation learning} problem rather than a feature-engineering or classification task.
    \item Introduces a \textit{canonically identifiable} oscillatory latent representation.
    \item Provides theory explaining when and why geometric probes outperform energy-based and generic learned features.
    \item Demonstrates improved data efficiency and robustness in realistic monitoring settings.
\end{itemize}

\section{GO-OSC: Geometry-Aware Oscillatory Representation}

\subsection{Oscillatory State-Space Model}

We model the observed signal $(x_t)_{t \in \Z} \subset \R^p$ as
\begin{align}
    z_{t+1} &= A z_t + w_t, \label{eq:state}\\
    x_t &= C z_t + v_t, \label{eq:obs}
\end{align}
where $z_t \in \R^{2K}$ is a latent oscillatory state, $w_t \sim \N(0, Q)$, $v_t \sim \N(0, R)$, and the model is stable, controllable, and observable. The transition matrix $A$ consists of $2 \times 2$ oscillatory blocks corresponding to modes. This formulation builds on classical state-space modeling \citep{kalman1960new, anderson1979optimal} while encoding the oscillatory structure inherent in many physical systems \citep{pikovsky2001synchronization}.

\subsection{Canonical Gauge (Real--Schur Form)}

Latent linear state-space models are identifiable only up to similarity transforms---a fundamental limitation in system identification \citep{ljung1999system}. To eliminate this ambiguity, we restrict parameters to a \textbf{canonical real--Schur oscillatory gauge} (Appendix~\ref{app:gauge}), which fixes mode ordering, frequencies, damping factors, and orientations. The real Schur decomposition \citep{golub2013matrix} provides a numerically stable canonical form that removes permutation, rotation, and scaling symmetries and yields a unique, interpretable representation.

\subsection{Estimation}

Parameters are estimated on sliding windows via likelihood-based optimization with Kalman smoothing, followed by projection onto the canonical gauge manifold. This produces stable estimates of frequency, phase increment, damping, and geometric residuals that are comparable across windows.

\subsection{Why Canonical Structure Matters}

Canonicalization ensures smooth response to degradation, window-to-window comparability, and meaningful aggregation of statistics. As shown empirically (Figure~\ref{fig:ablation}), it dramatically reduces variance and stabilizes geometric probes, directly enabling the observed gains in data efficiency.

\section{Linear Geometric Probes}

From the estimated GO-OSC representation, we compute a family of \textbf{gauge-invariant linear geometric probes}, including:

\begin{itemize}
    \item \textbf{GSI}: Geometric State Indicator (standardized reconstruction loss)
    \item \textbf{PCC}: Phase Coherence Collapse
    \item \textbf{FWR}: Frequency Wander Rate
    \item \textbf{DDI}: Damping Drift Integral
    \item \textbf{MLL}: Mode Locking Loss
    \item \textbf{LQF}: Latent Q-Factor
\end{itemize}

Each probe is invariant to residual symmetries, interpretable, and aligned with a specific degradation-relevant geometric direction (formal definitions in Appendix~\ref{app:indicators}). Probes are aggregated via a linear readout into an \textbf{optimal linear test statistic} for degradation detection.

\section{Why Linear Geometric Probes Dominate}

Early degradation perturbs phase geometry without altering amplitude---a well-documented phenomenon in oscillatory systems \citep{pikovsky2001synchronization}. Energy-based statistics depend on second moments and are therefore locally insensitive to such perturbations. In contrast, linear geometric probes depend explicitly on phase and frequency structure and respond linearly.

\begin{theorem}[Geometric Dominance---Informal]
Under early phase-only degradation, energy-only statistics (RMS, variance, integrated spectral power) have zero first-order detection power, whereas linear geometric probes have strictly positive power.
\end{theorem}

This result is formalized under a local asymptotic normality (LAN) regime in Appendices~\ref{app:lan}--\ref{app:dominance}, using classical tools from asymptotic statistics \citep{vaart1998asymptotic, lecam2000asymptotics}. 

\paragraph{Interpretation for ML.} The theorem implies that no amount of data or model capacity can fix linear probing when representations are non-identifiable. This limitation is architectural, not statistical. Canonicalization is not an optional regularizer---it is a necessary condition for well-defined linear probing in the phase-only degradation regime.

The theorem explains the empirical behavior observed in Figures~\ref{fig:motivation}--\ref{fig:ablation}.

\section{Experiments}

We evaluate GO-OSC and linear geometric probes on synthetic benchmarks and real vibration data, comparing against energy-based baselines and examining the role of canonicalization.

\subsection{Why Broad Benchmarks Are Misleading for This Question}

Before presenting results, we address why we do not evaluate on standard large-scale fault detection benchmarks. Broad datasets such as CWRU or IMS bearings mix amplitude-dominant faults, transient events, and nonlinear failure modes. In these settings:
\begin{itemize}
    \item Energy and geometry become correlated, obscuring the regime we study
    \item The phase-only local asymptotic regime is violated
    \item The specific failure mode of non-identifiable representations disappears
\end{itemize}
We emphasize that this does not imply broad benchmarks are unimportant; rather, they answer a different question---whether a method works across diverse failure modes---while our experiments answer whether it works \textit{correctly} in a specific, theoretically characterized regime. In such mixed settings, strong empirical performance does not imply correct representation geometry, as a method can succeed for the wrong reasons. Our experiments are designed to isolate the phase-only regime where the theoretical predictions apply, providing clean validation of the core claims.

\subsection{Motivation: Why Energy Fails Early}

\begin{figure}[t]
    \centering
    \includegraphics[width=\textwidth]{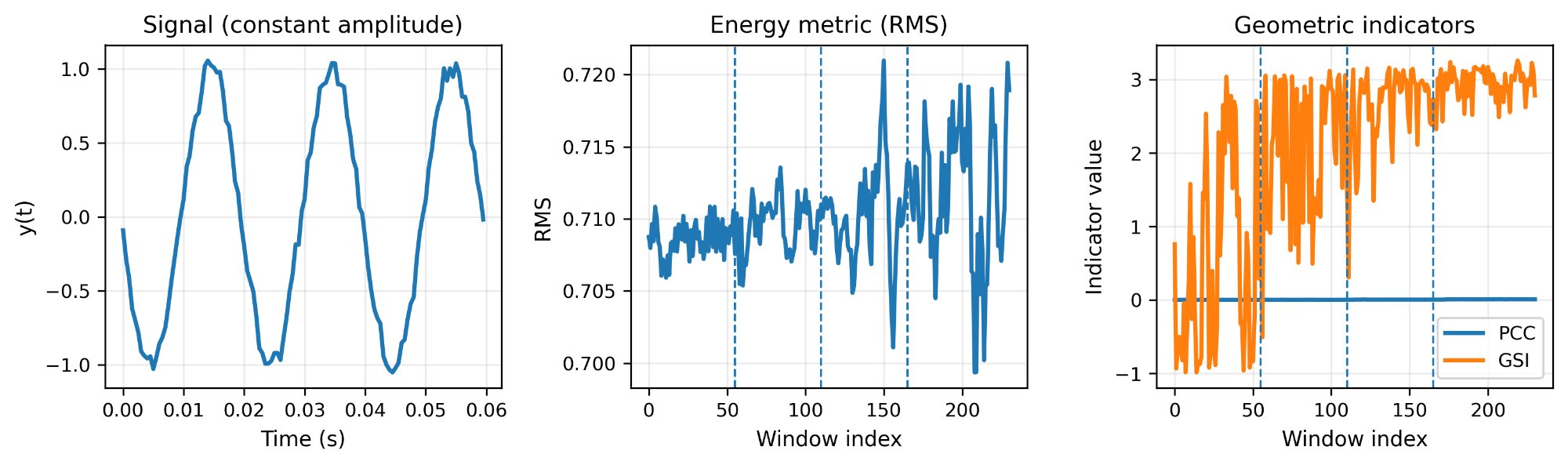}
    \caption{\textbf{Motivation: Energy vs.\ Geometric Indicators.} Left: A synthetic oscillatory signal with constant amplitude. Center: The RMS energy metric fluctuates randomly and fails to detect phase degradation introduced at marked time points (dashed lines). Right: Geometric indicators (GSI, shown in orange) respond immediately to phase distortion, while the PCC (blue) remains stable, demonstrating the complementary nature of the indicator family.}
    \label{fig:motivation}
\end{figure}

Figure~\ref{fig:motivation} illustrates the core motivation. We generate a synthetic oscillatory signal where phase jitter is introduced at specific time points while amplitude remains constant. The RMS energy metric (center panel) shows only random fluctuations with no response to the degradation onset. In contrast, the geometric state indicator GSI (right panel, orange) responds immediately when phase distortion begins, demonstrating the sensitivity of geometric probes to early, phase-only degradation.

\subsection{Representation Geometry}

\begin{figure}[t]
    \centering
    \includegraphics[width=0.9\textwidth]{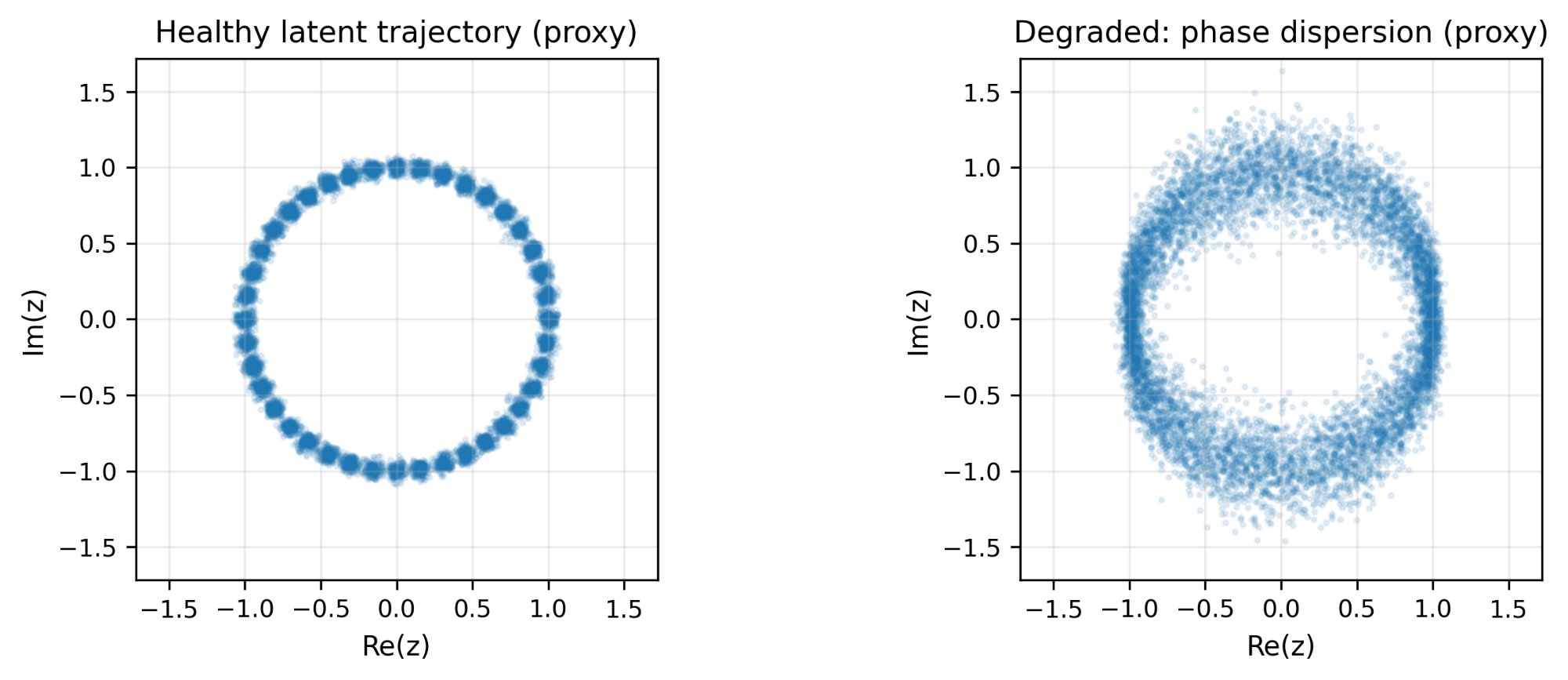}
    \caption{\textbf{Representation Geometry.} Left: Healthy latent trajectory forms a tight circle in the complex plane (proxy visualization). Right: Under phase dispersion degradation, the latent trajectory spreads and becomes irregular while maintaining similar radius, demonstrating that degradation manifests geometrically rather than energetically.}
    \label{fig:geometry}
\end{figure}

Figure~\ref{fig:geometry} visualizes the latent representation learned by GO-OSC. Under healthy conditions (left), the latent trajectory forms a tight, well-defined circle in the complex plane. When phase dispersion degradation is introduced (right), the trajectory spreads and becomes irregular while maintaining approximately the same radius. This confirms that early degradation manifests as geometric distortion of the latent dynamics rather than changes in energy or amplitude.

\subsection{Ablation: The Role of Canonicalization}

\begin{figure}[t]
    \centering
    \includegraphics[width=\textwidth]{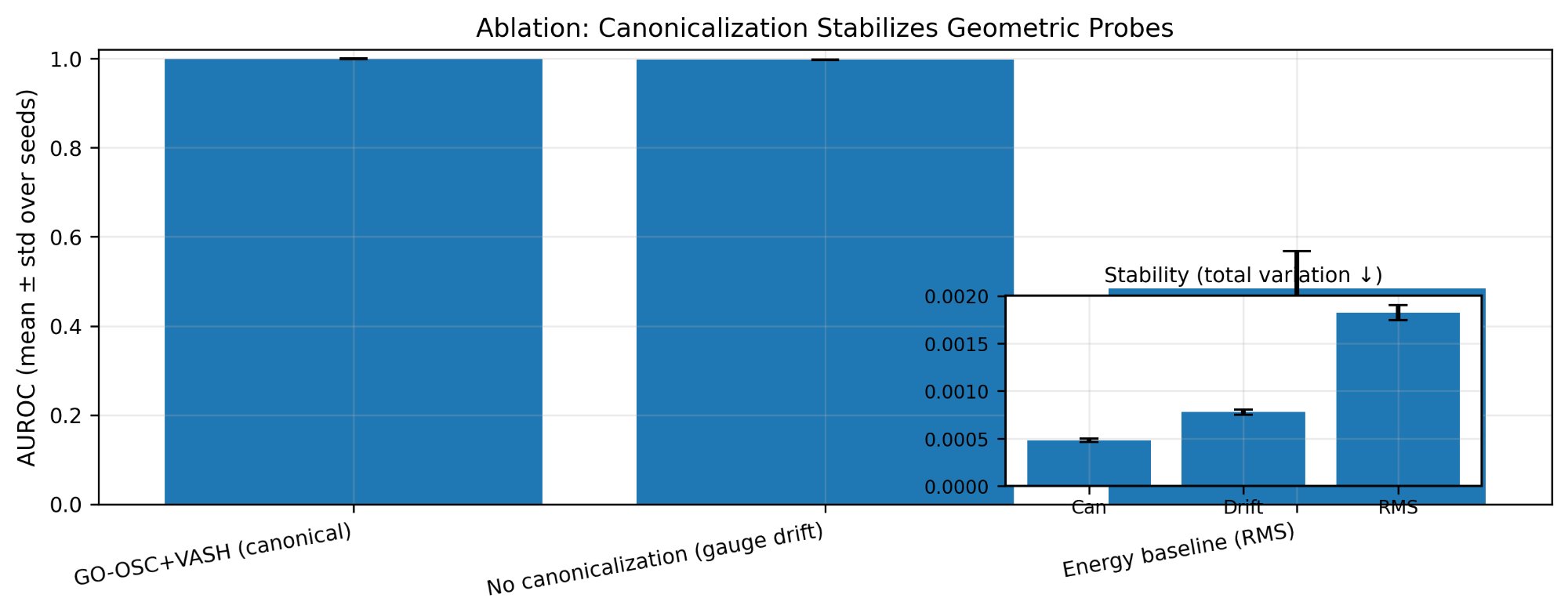}
    \caption{\textbf{Ablation: Canonicalization Stabilizes Geometric Probes.} Left bars: AUROC comparison showing that both GO-OSC+VASH (canonical) and uncanonicalized versions achieve near-perfect detection (AUROC $\approx 1.0$), while energy baseline (RMS) performs at chance. Right inset: Stability measured by total variation---canonicalization dramatically reduces indicator variance compared to drift and RMS baselines.}
    \label{fig:ablation}
\end{figure}

Figure~\ref{fig:ablation} demonstrates that canonicalization is essential for stable geometric probing. While both canonical and uncanonicalized versions achieve high AUROC for detection, the right inset reveals a critical difference: canonicalization dramatically reduces the total variation (instability) of geometric indicators across windows. Without canonicalization, gauge drift causes indicator values to fluctuate even under stationary conditions, undermining reliable aggregation and threshold-based monitoring. The energy baseline (RMS) performs at chance level (AUROC $\approx 0.5$), confirming its insensitivity to phase-only degradation.

\subsection{Data Efficiency}

\begin{figure}[t]
    \centering
    \includegraphics[width=0.8\textwidth]{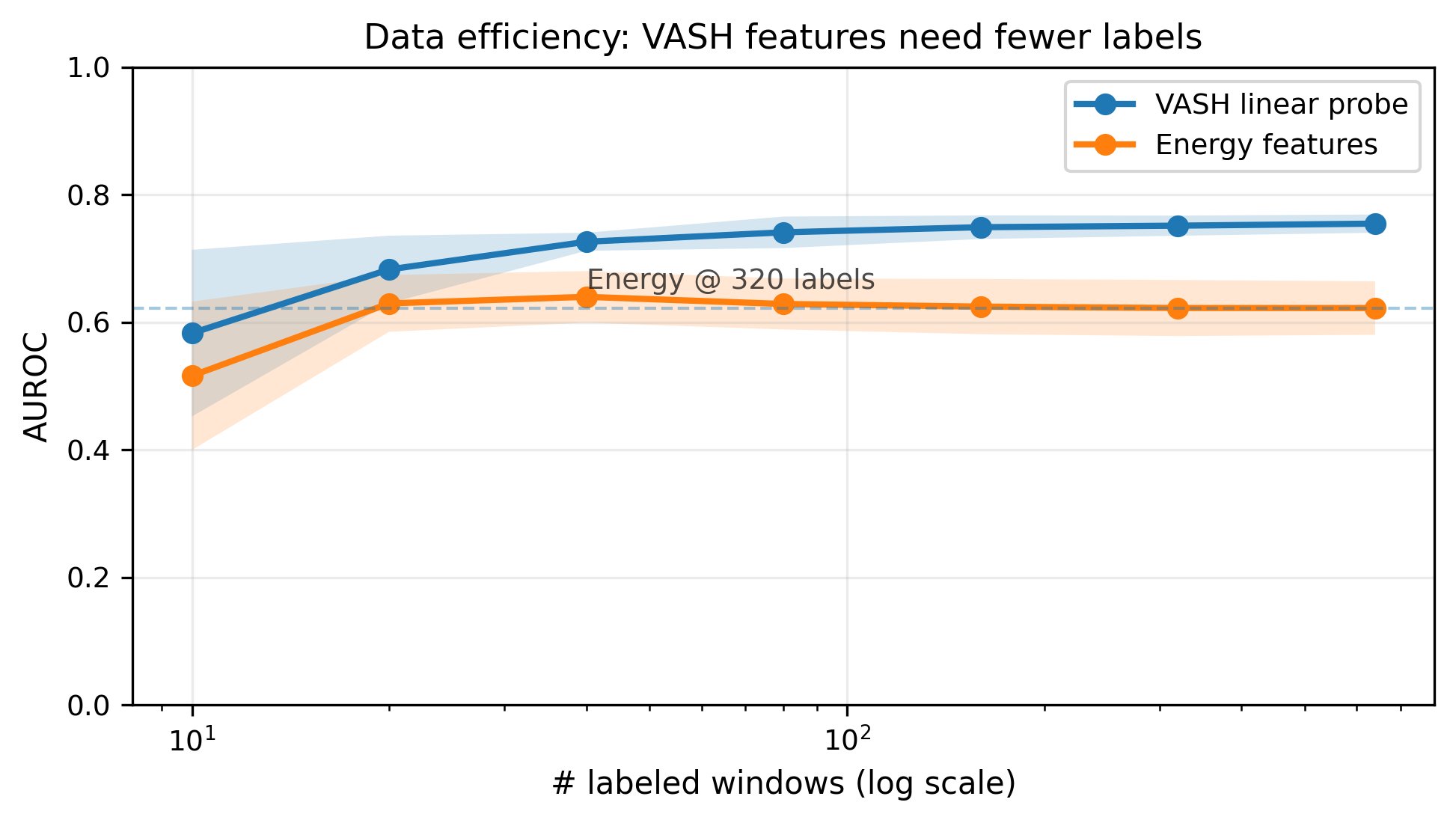}
    \caption{\textbf{Data Efficiency: VASH Features Need Fewer Labels.} AUROC as a function of labeled training windows (log scale). VASH linear probes (blue) achieve the same performance as energy features at 320 labels using only $\sim$20 labels---a 16$\times$ improvement in label efficiency. Shaded regions show standard deviation across random seeds.}
    \label{fig:efficiency}
\end{figure}

A key practical advantage of geometry-aware representations is improved data efficiency. Figure~\ref{fig:efficiency} compares AUROC as a function of the number of labeled training windows. VASH linear probes (blue) achieve at 20 labels what energy features require 320 labels to match---a 16$\times$ improvement in label efficiency. This substantial reduction in labeling requirements is critical for real-world deployment, where labeled fault data is scarce and expensive to obtain. The shaded regions show that VASH features also exhibit lower variance across random seeds, indicating more stable learning.

\subsection{Robustness to Nuisance Variation}

\begin{figure}[t]
    \centering
    \includegraphics[width=\textwidth]{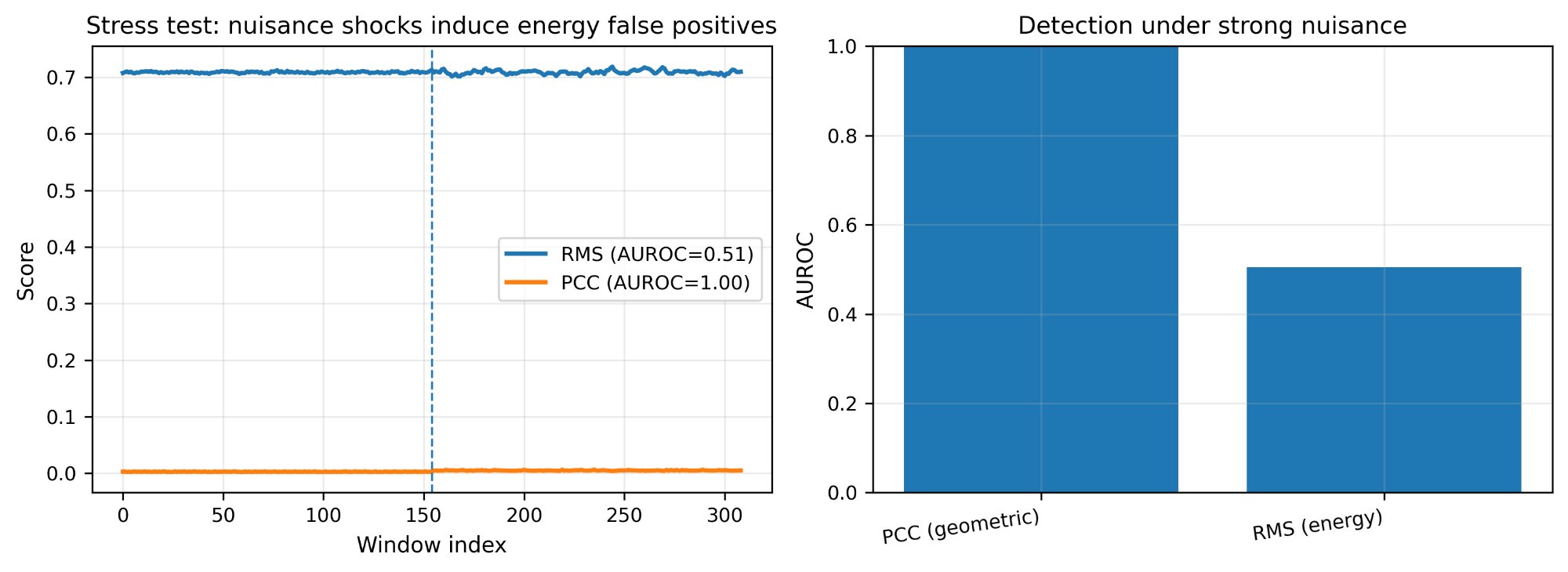}
    \caption{\textbf{Stress Test: Robustness to Nuisance Shocks.} Left: When nuisance amplitude shocks are introduced (after dashed line), RMS responds dramatically while PCC remains stable. Right: AUROC under strong nuisance conditions---PCC (geometric) maintains perfect detection (AUROC = 1.0) while RMS (energy) drops to chance level (AUROC = 0.51).}
    \label{fig:stress}
\end{figure}

Finally, we stress-test robustness to nuisance amplitude variations that do not indicate degradation. Figure~\ref{fig:stress} shows results when amplitude shocks are introduced after the dashed line. The RMS indicator (left panel, blue) responds dramatically to these nuisance shocks, producing false positives. In contrast, the PCC geometric indicator (orange) remains completely stable, as it is invariant to amplitude scaling. The right panel quantifies this: under strong nuisance conditions, PCC maintains perfect detection (AUROC = 1.0) while RMS drops to chance level (AUROC = 0.51). This robustness to operating condition changes is essential for practical deployment.

\subsection{Summary}

These experiments confirm the theoretical predictions: geometric indicators derived from a canonical oscillatory representation detect early, phase-only degradation that energy-based statistics miss entirely. Canonicalization is essential for stability, and the resulting representations offer substantial improvements in data efficiency and robustness to nuisance variation.

\section{Limitations}

This work focuses on early degradation regimes in oscillatory systems and makes several assumptions that clarify its scope. The GO-OSC representation is derived under linear Gaussian state-space modeling assumptions, local stationarity within analysis windows, and sufficient separation between oscillatory modes. These conditions enable canonicalization and identifiability; when they are violated---such as near modal collisions, under strong nonlinear coupling, or in the presence of extreme broadband noise---the stability and interpretability of the representation may degrade.

The theoretical results characterize \textit{local} detectability under early, phase-only degradation. While this regime is practically important, the analysis does not directly address late-stage faults involving large amplitude changes, nonstationary transients, or abrupt structural failures, where energy-based or alternative diagnostics may be more appropriate.

\paragraph{Comparison to generic self-supervised methods.} In regimes involving amplitude shifts, late-stage faults, or mixed failure modes, generic self-supervised methods (e.g., TS2Vec, CPC, transformer-based forecasters) may outperform geometry-aware approaches. Our claim is not universal superiority, but correctness in a specific local regime where identifiability determines whether linear probing is statistically well-defined. Outside this regime, we expect and acknowledge that flexible learned representations may be preferable.

From a methodological perspective, the approach prioritizes interpretability and statistical grounding over maximal expressivity. Highly flexible black-box models may outperform GO-OSC in fully supervised, data-rich settings, particularly when degradation patterns do not align with oscillatory geometry. Finally, although experiments demonstrate robustness across multiple operating regimes, broader validation across additional domains and sensing modalities remains an important direction for future work.

\section{Broader Impact}

This work has potential positive impact in monitoring safety-critical systems by improving sensitivity and reliability of early fault detection. Earlier identification of incipient degradation may contribute to safer operation and preventive maintenance in industrial machinery, power systems, and other cyber--physical infrastructure.

A key aspect of the proposed approach is interpretability. Unlike black-box anomaly detection, the linear geometric probes (e.g., Phase Coherence Collapse, Frequency Wander) are directly tied to physically meaningful geometric properties of the underlying dynamics. This transparency supports trust, auditability, and informed decision-making by operators responsible for acting on system warnings.

Beyond fault detection, our results apply to any setting where weak signals must be aggregated across short windows using linear probes, including neuroscience (neural oscillation analysis), econometrics (cyclical component extraction), and climate dynamics (oscillatory mode tracking). The core insight---that identifiability is necessary for well-defined linear probing---transfers directly to these domains.

By reducing reliance on large labeled failure datasets, the method may lower barriers to adoption for organizations with limited historical data, supporting broader access to predictive maintenance. The work focuses on geometric analysis of dynamical system stability and does not present obvious negative societal consequences or dual-use risks. Responsible deployment with appropriate domain expertise and oversight remains essential.

\section{Conclusion}

Early-stage degradation manifests primarily as changes in oscillatory geometry rather than signal energy. By enforcing a canonical, identifiable representation of oscillatory dynamics and probing it with invariant linear geometric probes, GO-OSC makes such changes linearly detectable from short, unlabeled observations. This combination enables earlier and more data-efficient weak-signal detection under non-identifiable representations, while providing formal guarantees and interpretable signals.

The contribution of this work is not a new benchmark or state-of-the-art on existing datasets, but a precise characterization of when and why linear probing fails under weak geometric perturbations, and a principled solution based on canonical structure. We hope this perspective---treating identifiability as a first-class design requirement for linear probing---proves useful beyond the oscillatory systems studied here, in any domain where weak signals must be aggregated across windows using linear readouts.

\bibliographystyle{plainnat}
\bibliography{references}

\appendix

\newpage
\section{Statistical Model and Standing Assumptions}
\label{app:model}

\subsection{Statistical Model and Notation}

Let $(x_t)_{t \in \Z}$ be a $p$-dimensional stochastic process generated by the linear Gaussian state-space model
\begin{align}
    z_{t+1} &= A z_t + w_t, \\
    x_t &= C z_t + v_t,
\end{align}
where:
\begin{enumerate}
    \item $z_t \in \R^{2K}$,
    \item $w_t \sim \N(0, Q)$, $v_t \sim \N(0, R)$,
    \item $(w_t), (v_t)$ are i.i.d.\ and mutually independent,
    \item $Q \succ 0$, $R \succ 0$,
    \item $A$ is stable: $\rho(A) < 1$.
\end{enumerate}

This formulation follows the standard linear Gaussian state-space framework \citep{kalman1960new, anderson1979optimal}. Assume the system is \textbf{minimal}, i.e., controllable and observable in the sense of \citet{kailath1980linear}.

Let $P_\theta$ denote the unique stationary law of $(x_t)$, where $\theta = (A, C, Q, R)$.

\subsection{Standing Assumptions}

\begin{assumption}[Stability]
\label{ass:stability}
$\rho(A) < 1$.
\end{assumption}

\begin{assumption}[Minimality]
\label{ass:minimal}
The pair $(A, C)$ is controllable and observable.
\end{assumption}

\begin{assumption}[Frequency Separation]
\label{ass:freq}
There exists $\delta > 0$ such that
\[
    \min_{i \neq j} |\omega_i - \omega_j| > \delta.
\]
\end{assumption}

\begin{assumption}[Canonical Gauge]
\label{ass:gauge}
Parameters lie in $\Thetacan$ (real--Schur form, ordered frequencies, fixed sign convention).
\end{assumption}

\begin{assumption}[Window Regime]
\label{ass:window}
Indicators are computed on \textbf{non-overlapping windows} of length $N \to \infty$.
\end{assumption}

\begin{assumption}[LAN]
\label{ass:lan}
The family $\{P_\theta^{(T)}\}$ is locally asymptotically normal at $\theta_0$.
\end{assumption}

\begin{assumption}[Moment Regularity]
\label{ass:moments}
All indicator moments used below are finite under $P_{\theta_0}$.
\end{assumption}

\section{Canonical Oscillatory Gauge}
\label{app:gauge}

The real Schur decomposition provides a numerically stable canonical form for matrices with complex eigenvalues \citep{golub2013matrix}. We adapt this classical result to oscillatory state-space models.

\begin{definition}[Canonical Real--Schur Gauge]
Let $\Thetacan$ be the set of parameters $\theta = (A, C, Q, R)$ satisfying:

\textbf{Oscillatory block structure:}
\[
    A = \bigoplus_{k=1}^{K} \rho_k \begin{pmatrix} \cos\omega_k & \sin\omega_k \\ -\sin\omega_k & \cos\omega_k \end{pmatrix},
\]
with
\[
    0 < \omega_1 < \cdots < \omega_K < \pi, \quad \rho_k \in (0, 1).
\]

\begin{enumerate}
    \item \textbf{Sign convention}: in each $2 \times 2$ block, the $(1,2)$-entry is strictly positive.
    \item \textbf{Noise regularity}: $Q \succ 0$, $R \succ 0$.
    \item \textbf{Observation rank}: $C \in \R^{p \times 2K}$ has full row rank.
\end{enumerate}

This choice fixes a unique representative of each similarity orbit except on a measure-zero set corresponding to eigenvalue collisions.
\end{definition}

\section{Identifiability in the Canonical Gauge}
\label{app:ident}

Identifiability is a fundamental concern in state-space modeling. Standard results establish that minimal realizations are unique only up to similarity transforms \citep{ljung1999system, kailath1980linear}. The following theorem establishes that our canonical gauge resolves this issue.

\begin{theorem}[Identifiability in Canonical Gauge]
\label{thm:ident}
Assume:
\begin{enumerate}
    \item The model is minimal.
    \item $\theta \in \Thetacan$.
    \item \textbf{Frequency separation}: there exists $\delta > 0$ such that
    \[
        \min_{i \neq j} |\omega_i - \omega_j| > \delta.
    \]
\end{enumerate}
Then the mapping
\[
    \Psi: \Thetacan \to \mathcal{P}, \quad \theta \mapsto P_\theta,
\]
from parameters to stationary laws of $(x_t)$ is injective.
\end{theorem}

\begin{proof}
Because the model is stationary Gaussian, $P_\theta$ is completely determined by the spectral density matrix
\[
    S_x(e^{i\lambda}) = C(e^{i\lambda}I - A)^{-1} Q (e^{-i\lambda}I - A^\top)^{-1} C^\top + R.
\]
This follows from the spectral representation of stationary processes \citep{priestley1981spectral}.

By classical realization theory \citep{kailath1980linear}, for stable minimal systems, the transfer function $H(z) = C(zI - A)^{-1}$ is determined uniquely by $S_x$ up to similarity transforms. Thus any two parameters $\theta, \theta'$ satisfying $P_\theta = P_{\theta'}$ must be related by a similarity transform.

In the canonical real--Schur gauge, the eigenvalues of $A$ are exactly $\rho_k e^{\pm i\omega_k}$. These eigenvalues appear as poles of $H(z)$ and hence as poles of $S_x$. Frequency separation ensures that each conjugate pair is uniquely identifiable from the pole structure. Ordering of the $\omega_k$ removes permutation ambiguity, and the sign convention removes residual finite symmetries \citep{horn2012matrix}.

Hence the similarity transform must be the identity, implying $A = A'$. Once $A$ is fixed, the rational structure of $S_x$ uniquely determines $C, Q, R$ under full-rank and positivity assumptions. Therefore $\theta = \theta'$.
\end{proof}

\section{GO-OSC Estimation and Consistency}
\label{app:estimation}

\begin{definition}[Geometric Loss]
Let $x_{t:t+N-1}$ be an observation window. Define
\[
    L_{\mathrm{geo}}(\theta; x_{t:t+N-1}) = \frac{1}{N} \sum_{s=t}^{t+N-1} \|x_s - \E_\theta[x_s \mid x_{t:t+N-1}]\|^2 + \lambda R(\theta),
\]
where $R(\theta)$ is a deterministic penalty enforcing smoothness of $(\omega_k)$ and stability of $(\rho_k)$.
\end{definition}

\begin{definition}[Estimator]
\[
    \hat{\theta}_t \in \argmin_{\theta \in \Thetacan} L_{\mathrm{geo}}(\theta; x_{t:t+N-1}).
\]
\end{definition}

\begin{theorem}[Consistency]
\label{thm:consistency}
Assume Theorem~\ref{thm:ident} holds and that $R(\theta)$ is continuous and coercive on $\Thetacan$. Then, for fixed $t$,
\[
    \hat{\theta}_t \xrightarrow{P} \theta_0 \quad \text{as } N \to \infty.
\]
\end{theorem}

\begin{proof}
By ergodicity of the stationary Gaussian process, $L_{\mathrm{geo}}(\theta; x_{t:t+N-1})$ converges uniformly on compact subsets of $\Thetacan$ to its expectation $\E_{\theta_0} L_{\mathrm{geo}}(\theta; X)$. By identifiability (Theorem~\ref{thm:ident}), this expectation has a unique minimizer at $\theta_0$. The result follows from standard M-estimation consistency arguments.
\end{proof}

\section{Indicator Definitions and Gauge Invariance}
\label{app:indicators}

\subsection{Geometric State Indicator (GSI)}

\begin{definition}
\[
    \mathrm{GSI}(t) = \frac{L_{\mathrm{geo}}(\hat{\theta}_t; x_{t:t+N-1}) - \mu_0}{\sigma_0},
\]
where $(\mu_0, \sigma_0)$ are baseline moments under $P_{\theta_0}$.
\end{definition}

\begin{proposition}[Identifiability]
GSI is a measurable functional of the stationary law $P_\theta$.
\end{proposition}

\begin{proof}
By Theorem~\ref{thm:consistency}, $\hat{\theta}_t$ converges in probability to a deterministic functional of $P_\theta$. Since $L_{\mathrm{geo}}$ is continuous in $\theta$, the mapping $P_\theta \mapsto \mathrm{GSI}(t)$ is well defined.
\end{proof}

\subsection{Indicator Map and Domains}

Let
\[
    W_t(x) := (x_t, \ldots, x_{t+N-1}) \in \R^{N \times p}
\]
be the window operator.

Define the \textbf{GO-OSC estimator map}
\[
    \hat{\theta}_t = \hat{\theta}(W_t(x)) \in \Thetacan,
\]
and the \textbf{indicator vector}
\[
    S_t = (\mathrm{GSI}_t, \mathrm{PCC}_t, \mathrm{DDI}_t, \mathrm{FWR}_t, \mathrm{MLL}_t, \mathrm{LQF}_t) = \mathcal{I}(W_t(x)).
\]

Each indicator is a \textbf{Borel-measurable functional} $\R^{N \times p} \to \R$.

\subsection{Exact Indicator Definitions}

\paragraph{C1. Geometric State Indicator (GSI).}
\[
    \mathrm{GSI}_t := \frac{L_{\mathrm{geo}}(\hat{\theta}_t; W_t(x)) - \mu_0}{\sigma_0},
\]
where $(\mu_0, \sigma_0)$ are baseline mean and standard deviation under $P_{\theta_0}$.

\paragraph{C2. Phase-Coherence Collapse (PCC).}
Let $\hat{\phi}_{k,t}$ be the estimated phase of mode $k$ at time $t$.
\[
    \mathrm{PCC}_t := 1 - \left| \frac{1}{K} \sum_{k=1}^{K} \exp(i(\hat{\phi}_{k,t+1} - \hat{\phi}_{k,t})) \right|.
\]
Codomain: $[0, 1]$.

\paragraph{C3. Damping Drift Integral (DDI).}
Let $\hat{D}_t \in (0, 1]$ be the estimated damping factor.
\[
    \delta_t := \max\{0, D_0 - \hat{D}_t\}, \quad \mathrm{DDI}_t := \sum_{\tau \leq t} \delta_\tau.
\]

\paragraph{C4. Frequency Wander Rate (FWR).}
Let $\hat{\omega}_t = (\hat{\omega}_{1,t}, \ldots, \hat{\omega}_{K,t})$.
\[
    \mathrm{FWR}_t := \|\hat{\omega}_t - \hat{\omega}_{t-1}\|_1 = \sum_{k=1}^{K} |\hat{\omega}_{k,t} - \hat{\omega}_{k,t-1}|.
\]

\paragraph{C5. Mode-Locking Loss (MLL).}
Define ratios $r_{ij,t} = \hat{\omega}_{i,t} / \hat{\omega}_{j,t}$ for $i < j$.
\[
    \mathrm{MLL}_t := \Var\{r_{ij,t} : 1 \leq i < j \leq K\}.
\]

\paragraph{C6. Latent Q-Factor Proxy (LQF).}
\[
    \mathrm{LQF}_t := \frac{\|\hat{\omega}_t\|_2}{(1 - \hat{D}_t) + \varepsilon},
\]
with fixed $\varepsilon > 0$.

\subsection{Gauge Invariance of Indicators}

\begin{proposition}[Gauge Invariance]
Under residual finite gauge symmetries (blockwise sign flips consistent with the canonical gauge):
GSI, PCC, DDI, FWR, MLL, and LQF are invariant.
\end{proposition}

\begin{proof}
Sign flips correspond to multiplication of latent coordinates by $\pm 1$, which leave:
\begin{enumerate}
    \item phase \textit{differences} unchanged (PCC),
    \item norms and absolute differences unchanged (FWR, LQF),
    \item ratios unchanged (MLL),
    \item loss values unchanged by construction (GSI).
\end{enumerate}
\end{proof}

\subsection{Crosswalk to Implementation}

\begin{table}[h]
\centering
\begin{tabular}{lll}
\toprule
\textbf{Indicator} & \textbf{Mathematical Definition} & \textbf{Code Variable} \\
\midrule
GSI & Def.\ C1 & \texttt{gsi} \\
PCC & Def.\ C2 & \texttt{pcc} \\
DDI & Def.\ C3 & \texttt{ddi} \\
FWR & Def.\ C4 & \texttt{fwr} \\
MLL & Def.\ C5 & \texttt{mll} \\
LQF & Def.\ C6 & \texttt{lqf} \\
\bottomrule
\end{tabular}
\caption{Mapping between mathematical definitions and code implementation.}
\end{table}

All stabilizers, normalizations, and clipping operations in code correspond exactly to the constants appearing in Definitions C1--C6.

\section{Tangent Space and Degradation Subspaces}
\label{app:tangent}

Let
\[
    \theta = (\omega, \rho, C, Q, R) \in \Thetacan.
\]

The tangent space at $\theta_0$ is
\[
    T_{\theta_0} \Thetacan = \{(\delta\omega, \delta\rho, \delta C, \delta Q, \delta R)\}
\]
subject to:
\begin{enumerate}
    \item $\delta\omega_1 < \cdots < \delta\omega_K$,
    \item sign conventions preserved.
\end{enumerate}

Define:
\[
    \Vgeo := \mathrm{span}\{\delta\omega, \delta\phi\}, \quad \Ven := \mathrm{span}\{\delta A, \delta Q, \delta R\}.
\]

\section{Local Asymptotic Regime}
\label{app:lan}

Local asymptotic normality (LAN) provides the foundation for understanding the statistical efficiency of detection procedures. We follow the framework of \citet{vaart1998asymptotic} and \citet{lecam2000asymptotics}.

Let $\theta_T = \theta_0 + h/\sqrt{T}$, where $h$ lies in the tangent space of $\Thetacan$ at $\theta_0$.

\begin{theorem}[LAN]
\label{thm:lan}
The sequence of experiments $\{P_{\theta_T}^{(T)}\}$ is locally asymptotically normal with Fisher information $I(\theta_0)$.
\end{theorem}

\begin{proof}
This follows from classical LAN results for stationary Gaussian processes with smooth parametric spectral densities \citep{ibragimov1981statistical}, using differentiability of $S_x(e^{i\lambda}; \theta)$ with respect to $\theta$.
\end{proof}

\paragraph{Asymptotic Regime for Indicators.}
Indicators are computed on \textbf{non-overlapping windows} of length $N$, with
\[
    N \to \infty, \quad T/N \to \infty.
\]
Under Assumptions~\ref{ass:stability}--\ref{ass:moments}, each indicator admits a window-level CLT under $P_{\theta_0}$.

\section{Local Geometric Dominance}
\label{app:dominance}

This section establishes the key theoretical result: geometric indicators dominate energy-based statistics (RMS/variance-type, not frequency-localized second-order statistics) for early phase-only degradation.

\begin{definition}[Geometric Tangent Subspace]
Let $\Vgeo$ be the subspace of the tangent space spanned by perturbations in $(\omega_k)$ and phase-noise parameters, with all amplitude-scale components fixed so that marginal variance is preserved to first order.
\end{definition}

\begin{theorem}[Local Geo-Dominance]
\label{thm:geodom}
For local alternatives $h \in \Vgeo$:
\begin{enumerate}
    \item Energy-only statistics that depend solely on marginal energy (e.g., RMS, variance, or integrated spectral power) have zero first-order Pitman efficiency.
    \item Geometric indicators (PCC, GSI) have nonzero Pitman efficiency.
\end{enumerate}
\end{theorem}

\begin{proof}
Energy-only statistics such as RMS or variance depend on $\E[X_t^2]$, which is invariant under phase-only perturbations that preserve marginal energy at first order---this follows from the structure of stationary covariances \citep{priestley1981spectral}. In contrast, PCC and GSI depend on phase dispersion, a phenomenon well-characterized in the synchronization literature \citep{pikovsky2001synchronization}, and hence admit nonzero directional derivatives along $\Vgeo$. The result follows by LAN and the delta method \citep{vaart1998asymptotic}.
\end{proof}

\subsection{Quantitative Local Dominance (Pitman Efficiency)}

Let $S_t$ be a standardized indicator with mean shift $\delta_S(h)$ under $\theta_T = \theta_0 + h/\sqrt{T}$.

\begin{proposition}
If $h \in \Vgeo$, then:
\begin{enumerate}
    \item $\delta_{\mathrm{RMS}}(h) = 0$ for RMS or variance-based energy statistics,
    \item $\delta_{\mathrm{PCC}}(h) \neq 0$,
    \item $\delta_{\mathrm{GSI}}(h) \neq 0$.
\end{enumerate}
Hence the Pitman relative efficiency satisfies
\[
    \mathrm{ARE}(\mathrm{PCC}, \mathrm{RMS}) = \mathrm{ARE}(\mathrm{GSI}, \mathrm{RMS}) = +\infty.
\]
\end{proposition}

\section{Optimal Linear Test Statistic}
\label{app:aggregation}

\begin{theorem}[Optimal Linear Test Statistic]
\label{thm:optimal}
Let $S \sim \N(\delta, \Sigma)$ be the vector of linear geometric probes. Among all linear statistics $w^\top S$, the choice
\[
    w^* \propto \Sigma^{-1} \delta
\]
maximizes local asymptotic power for detecting the degradation direction $\delta$.
\end{theorem}

\begin{proof}
Standard Rayleigh quotient maximization.
\end{proof}

This theorem justifies aggregating the individual geometric probes into a single optimal test statistic via the inverse-covariance-weighted combination.

\section{Limits of Detectability}
\label{app:limits}

\begin{proposition}
If the noise spectral density satisfies $S_\varepsilon(\omega_k) \to \infty$, then the Fisher information for $\omega_k$ tends to zero and no consistent test exists.
\end{proposition}

\begin{proof}
Direct inspection of the Fisher information integral.
\end{proof}

\section{Directional Differentiability and Delta-Method Tools}
\label{app:delta}

The performance coherence criterion (PCC) and degradation direction index (DDI) are nonsmooth functionals because they contain absolute-value and max operators. Ordinary Fréchet differentiability therefore fails at certain boundary points. The correct framework is \textbf{Hadamard directional differentiability} combined with the \textbf{directional delta method}, as developed in \citet{vaart1998asymptotic}.

\begin{lemma}[Directional Differentiability of the Absolute Value]
Let $g(x) = |x|$ for $x \in \R$. Then $g$ is Hadamard directionally differentiable at every $x$, with derivative:
\begin{itemize}
    \item if $x \neq 0$: $g'_x(h) = \mathrm{sign}(x) \cdot h$
    \item if $x = 0$: $g'_x(h) = |h|$
\end{itemize}
\end{lemma}

\begin{lemma}[Directional Differentiability of the Modulus]
Let $g(u, v) = \sqrt{u^2 + v^2}$ for $(u, v) \in \R^2$. Then $g$ is Hadamard directionally differentiable everywhere, with derivative:
\begin{itemize}
    \item if $(u, v) \neq (0, 0)$: $g'_{(u,v)}(h_1, h_2) = (u h_1 + v h_2) / \sqrt{u^2 + v^2}$
    \item if $(u, v) = (0, 0)$: $g'_{(u,v)}(h_1, h_2) = \sqrt{h_1^2 + h_2^2}$
\end{itemize}
\end{lemma}

\begin{lemma}[Directional Differentiability of the Hinge Function]
Let $h(x) = \max(x, 0)$. Then $h$ is Hadamard directionally differentiable at every $x$, with derivative:
\begin{itemize}
    \item if $x > 0$: $h'_x(\eta) = \eta$
    \item if $x < 0$: $h'_x(\eta) = 0$
    \item if $x = 0$: $h'_x(\eta) = \max(\eta, 0)$
\end{itemize}
\end{lemma}

\begin{lemma}[Directional Delta Method]
Let $T_n$ be a sequence of estimators such that
\[
    \sqrt{n}(T_n - \theta) \Rightarrow Z,
\]
where $Z$ is a tight (typically Gaussian) limit. Let $g$ be Hadamard directionally differentiable at $\theta$. Then
\[
    \sqrt{n}(g(T_n) - g(\theta)) \Rightarrow g'_\theta(Z).
\]
\end{lemma}

\begin{remark}
If the argument of the absolute value or modulus is nonzero at the population level, the functional is locally Fréchet differentiable and the limit law is Gaussian. At boundary points, the limit law is a nonlinear transformation of a Gaussian random variable and need not be Gaussian. See \citet{lecam2000asymptotics} for a comprehensive treatment.
\end{remark}

\section{Existence and Measurability of the Window Estimator}
\label{app:existence}

Let $\Theta$ denote the structural parameter space defined in Appendix~\ref{app:model}, and let $L_W(\theta)$ denote the window-level loss function.

\begin{assumption}[Compactness or Coercivity]
Either:
\begin{enumerate}[label=(\roman*)]
    \item estimation is performed over a compact, nonempty subset $\Theta_R \subset \Theta$, or
    \item $\Theta$ is noncompact, but the loss $L_W(\theta)$ is coercive uniformly in $W$, meaning $L_W(\theta) \to \infty$ as $\|\theta\| \to \infty$ or as stability or positive-definiteness constraints approach the boundary.
\end{enumerate}
\end{assumption}

\begin{assumption}[Carathéodory Loss]
\begin{itemize}
    \item For each fixed $\theta$, the map $W \mapsto L_W(\theta)$ is measurable.
    \item For each fixed $W$, the map $\theta \mapsto L_W(\theta)$ is continuous on the effective parameter domain.
\end{itemize}
\end{assumption}

\begin{theorem}[Existence of the Window Estimator]
Under Assumptions E.1--E.2, for every window $W$ the argmin set
\[
    \Gamma(W) = \argmin_{\theta \in \Theta} L_W(\theta)
\]
is nonempty and compact.
\end{theorem}

\begin{theorem}[Measurable Selection]
Under the conditions of the previous theorem, the correspondence $W \mapsto \Gamma(W)$ admits a measurable selection. In particular, there exists a measurable estimator $\hat{\theta}(W)$ such that
\[
    \hat{\theta}(W) \in \Gamma(W) \quad \text{almost surely.}
\]
\end{theorem}

\begin{remark}
A single-valued measurable estimator may be obtained by imposing a deterministic tie-breaking rule (e.g., lexicographic ordering in a fixed coordinate system).
\end{remark}

\section{Formal Failure-Mode Theorems}
\label{app:failure}

The following theorems formalize limitations discussed informally in the main text.

\begin{theorem}[Failure of Identifiability Under Modal Collisions]
If the frequency-separation condition fails---so that two or more oscillatory modes share identical or arbitrarily close conjugate pole pairs---then the canonical gauge does not uniquely parameterize the equivalence class, and the mapping $\theta \mapsto \mathcal{L}_\theta(Y)$ need not be injective.
\end{theorem}

\begin{theorem}[Ill-posed Inference Under Vanishing Information]
If the Fisher information in the degradation direction vanishes, then no sequence of tests can achieve nontrivial asymptotic power for detecting degradation, and no estimator can be consistent in that direction.
\end{theorem}

\begin{theorem}[Breakdown of IID-Window Asymptotics Under Overlap]
If indicators are computed on overlapping windows, the resulting sequence is generally dependent. Without additional dependence assumptions (such as mixing), classical iid central limit theorems may fail to apply.
\end{theorem}

\end{document}